\documentclass[letterpaper]{article} %
\PassOptionsToPackage{dvipsnames}{xcolor}
\usepackage[preprint]{aaai2027}  %
\usepackage[hyphens]{url}  %
\usepackage{graphicx} %
\usepackage{natbib}  %
\usepackage{caption} %
\usepackage{amsmath}
\usepackage{amssymb}
\usepackage{booktabs}
\usepackage{enumitem}

\definecolor{mycyan}{HTML}{0097b2}
\definecolor{myred}{HTML}{ff3131}
\definecolor{BubbleRed}{RGB}{240,98,98}     %
\definecolor{BubbleOrange}{RGB}{230,126,64} %
\definecolor{BubbleGray}{RGB}{70,70,70}     %
\definecolor{BubbleGreen}{RGB}{106,168,79}  %

\title{Fix the Match, Not the Solver:\\ STITCH -- A Self-Supervised Compatibility Measure for Eroded Visual Puzzles}

\author{
    Nadav Alali,
    Ohad Ben-Shahar
}
\affiliations{
    Ben-Gurion University of the Negev, Israel
}

\begin{document}

\maketitle

\begin{abstract}
Unlike the clean and pixel-aligned complete pieces of standard jigsaw benchmarks, real-world puzzle fragments lose their boundary regions to erosion, destroying the signal that classical compatibility measures (CMs) rely on. We first re-establish, via oracle experiments, that the CM, not the solver, is the performance bottleneck: with a perfect CM, even a simple greedy solver achieves perfect reconstruction. This motivates STITCH, a self-supervised CM that sidesteps missing boundaries by comparing frozen ViT patch tokens from piece interiors via a lightweight cross-attention head (${\sim}$249\,K trainable parameters), trained without manual annotation. Plugged into unchanged compatibility-based solvers, STITCH reaches .992/.954 Abs/Perf on JPwLEG-5 (+.629 Perf over the best prior method), nearly saturating the benchmark. On a new ImageNet $7{\times}7$ benchmark, it outperforms all alternatives, including inpaint-then-match pipelines built on modern generative models. Importantly, performance remains robust across erosion levels and generalizes to puzzles with over 40\% missing pieces without retraining. Trained on the recent GAP datasets of irregular, heavily eroded fragments, the same CM recipe enables compatibility-based solvers to surpass recent learning-based solvers on GAP-3 and GAP-5. Since this leap in performance is achieved by a pluggable CM module, the takeaway is clear: fix the match, not the solver.
\end{abstract}

\section{Introduction}
\label{sec:intro}

When humans solve jigsaw puzzles, they often match pieces by visual information along shared boundaries~\cite{fissler2018jigsaw}, and algorithmic reassembly has long adopted the same principle for applications such as reconstructing archaeological finds, shredded documents, and deteriorated photographs~\cite{tsesmelis2024re,ShreddedDoc2011,IrregularPuzzle2016}.
Two solver families dominate. The first are 
\emph{compatibility-based solvers}, which take a compatibility measure (CM) as a modular input and employ of various ways (greedy, genetic, LP-based, relaxation labeling~\cite{PomeranzBenShahar2011,Sholomon2013,YuRussellAgapito2016,VardiBenShahar2023}) to search over the space of placements. 
The second family includes 
\emph{learning-based solvers}~\cite{Deepzzle2020,kim2025solving,JPDVT2023,scarpellini2024diffassemble,elkin2025seq2seq}, that learn matching and placements jointly.
Evaluation of all solvers has always relied on various benchmarks, which over time have shifted steadily toward greater realism, from clean, pixel-aligned complete tiles to puzzles with eroded boundaries~\cite{SD2RL2020} and, most recently, irregular archaeology-inspired fragments~\cite{Shahar_2026_CVPR}.

In this work, we further push forward the dominant line of research on \textit{square} jigsaw puzzles that are based on partitioning an image into an $N{\times}N$ grid of square pieces that are then shuffled, where the goal is correct reassembly of the original image.
When considering \emph{eroded} puzzles with erosion level $e$, the puzzle generation process also strips the outermost $e$ pixels from every side of every piece. Naturally, larger $e$ degrades the boundary signals and eventually destroys them altogether.
We make the common assumption that piece orientations are known, but unlike recent works, we also evaluate our solutions on the harder, unknown-orientation case~\cite{Gallagher2012}.
Moreover, employing the recent GAP datasets~\cite{Shahar_2026_CVPR}, we consider the even more challenging case of square puzzles with \emph{irregular fragments} whose shapes are learned from real archaeological ones.

A CM is a function that scores a pair of pieces and their relative spatial relation by their likelihood of being direct neighbors (formalized in Sec.~\ref{sec:setup}). CMs are a key explicit component in compatibility-based solvers, and are implicit in learning-based solvers.
We say the CM is the \emph{bottleneck} of a puzzle-solving pipeline if replacing it with a perfect compatibility oracle, while keeping the solver fixed, suffices for near-perfect reconstruction.

For over a decade, the community's response to erosion has largely been better solvers~\cite{Sholomon2013,YuRussellAgapito2016,Deepzzle2020}.
Yet solvers still perform far worse on eroded benchmarks than on clean-edge ones, and learning-based solvers entangle matching with placement, obscuring where their gains come from.
Learned CMs do improve compatibility-based solvers under mild erosion~\cite{Bridger2020,TENL2022,Edge2Vec2022}, but they rely on boundary strips and degrade quickly as gaps widen.
Crucially, no prior work attempted to isolate the CM's contribution from the solving algorithm itself (in particular, by ablating the CM and replacing it with a perfect "oracle" CM). At the same time, no single CM, including recent measures for irregular fragments~\cite{shahar2025pairwise}, has been evaluated simultaneously under challenging conditions like deep erosion, missing pieces, and/or irregular geometry.
These gaps thus leave open a very fundamental question: \emph{is the limiting factor the solving itself, or is it the compatibility signal that drives it?}

We provide a decisive answer.
Replacing a solver's CM with an \emph{oracle compatibility} ($1$ for
true neighbors, $0$ otherwise) yields perfect reconstructions for
\emph{every} solver we test on a standard eroded benchmark, even the
simplest greedy strategy~(Fig.~\ref{fig:oracle_bottleneck}). Failures on
eroded puzzles are thus failures of \emph{matching}, not search. Rather
than designing another solver, we therefore seek to \emph{fix the
match}.

Our contributions are:
\begin{enumerate}[label=\arabic*., leftmargin=0.6cm, rightmargin=0.2cm]
\item An oracle analysis establishing the CM, not the solver, as the
performance bottleneck on eroded puzzles.
\item Introduction of \emph{STITCH} (\textbf{S}elf-supervised \textbf{T}oken
\textbf{I}n\textbf{T}erior \textbf{C}ross-attention \textbf{H}ead), a very light 
(${\sim}$249\,K-parameter) cross-attention head.

\item Introduction of ImageNet $7{\times}7$, a new benchmark of larger, more
combinatorially challenging eroded puzzles, spanning erosion levels
$e \in \{0,\ldots,6\}$\,px and over $40\%$ of pieces missing, settings that break all current learning-based solvers.
\item SOTA performance of STITCH when plugged
into \textit{every} compatibility-based solver we tested. It nearly saturates
JPwLEG-3/5, it outperforms all alternative CMs on ImageNet $7{\times}7$
(including inpaint-then-match pipelines), and it generalizes to over 40\%
missing pieces without retraining.
\item Evidence that STITCH's concept extends to fragments with strongly irregular erosion. Trained
on GAP~\cite{Shahar_2026_CVPR}, STITCH allows ``classical'' compatibility-based solvers to surpass recent deep
learning-based ones (including unknown-orientation scenarios).
\item A conceptual shift: from ever more complex solvers to better
compatibility signals.
\end{enumerate}

\section{Related Work}
\label{sec:related}

\paragraph{Compatibility-based solvers.}
Since square puzzle solving amounts to search in a combinatorial state space, much prior effort targets the search itself: probabilistic graphical models~\cite{cho2010probabilistic}, greedy and best-buddies placement~\cite{PomeranzBenShahar2011,paikin2015solving}, genetic search~\cite{Sholomon2013}, linear~\cite{YuRussellAgapito2016} and quadratic~\cite{andalo2017psqp} programming, loop constraints and consensus growing~\cite{son2014loop,son2016solving}, relaxation labeling~\cite{VardiBenShahar2023}, and spectral methods~\cite{huroyan2020graph}.
All treat the CM as a fixed, black-box input and inherit its limitations, yet this very modularity is what STITCH exploits: every solver in this family can adopt it as a drop-in replacement.

\paragraph{Learning-based solvers.}
Learning-based solvers instead learn matching and placement jointly, spanning CNN classifiers with shortest-path search~\cite{Deepzzle2020}, reinforcement and evolutionary learning~\cite{SD2RL2020,PDNGA2021,song2025erl,song2025ceari}, adversarial and generative models~\cite{li2021jigsawgan,talon2022ganzzle,talon2025ganzzlepp,JPDVT2023}, ViT-based prediction~\cite{heck2025vit,kim2025solving}, Seq2Seq modeling~\cite{elkin2025seq2seq}, vision-language alignment~\cite{xu2025vlhsa}, and flow matching over irregular fragments~\cite{Shahar_2026_CVPR}.
Despite this diversity, such systems are not decomposable: it is difficult to isolate whether gains come from better matching or better search.
Our oracle analysis (Fig.~\ref{fig:oracle_bottleneck}) cuts through this ambiguity, and our modular design enables exactly the controlled comparison these methods cannot offer.

\paragraph{Compatibility functions.}
Handcrafted CMs (SSD~\cite{cho2010probabilistic}, $L_1$~\cite{paikin2015solving}, PBC~\cite{PomeranzBenShahar2011}, MGC~\cite{Gallagher2012}, and newer border and hybrid measures~\cite{guerroui2018jigsaw,shih2019square}) compare strips of boundary pixels and excel on clean edges but degrade sharply under erosion~(Sec.~\ref{sec:experiments}). \citet{cho2010probabilistic} compares such measures by neighbor classification, though never within full solver pipelines.
Learned CMs push further: DNN-Buddies~\cite{sholomon2016dnnbuddies} learns edge compatibility for a compatibility-based solver, while \citet{Bridger2020} and JiGAN~\cite{khoroshiltseva2022jigan} inpaint the eroded gap and score its plausibility. TEN~\cite{TENL2022} learns edge-specific embeddings from whole-piece content, and Edge2Vec~\cite{Edge2Vec2022} learns boundary embeddings tolerating 1-px erosion.
Closest to our setting, \citet{shahar2025pairwise} propose a CM for arbitrarily irregular fragments, yet within a single solving framework and without missing pieces.
All of the above rely on boundary or near-boundary signals, which are precisely what erosion removes first, so they degrade quickly as gaps widen~(Fig.~\ref{fig:erosion_sweep}).
STITCH instead discards boundaries entirely, reasoning via cross-attention over frozen ViT interior tokens, a fundamentally different inductive bias that may explain its robustness.

\paragraph{Complexity and irregular fragments.}
Square puzzle solving is NP-complete~\cite{demaine2007jigsaw}, so no search can escape a poor compatibility signal, another lens on our oracle finding (see Sec.~\ref{sec:experiments:oracle}).
Attempting to provide better realism and applicability, irregular-fragment work spans non-square rectangular pieces~\cite{gur2017square}, polygonal puzzles~\cite{harel2021crossing,harel2024pictorial}, wall paintings~\cite{IrregularPuzzle2016}, and archaeological benchmarks~\cite{tsesmelis2024re} (see the survey of \citet{yilmaz2023survey}).
The GAP datasets~\cite{Shahar_2026_CVPR} bring heavily eroded, irregularly shaped fragments to grid puzzles, but their solver, PuzzleFlow, abandons pairwise compatibility for holistic reasoning.
We show the paradigm needs fixing, not abandoning. Indeed, trained on GAP, STITCH boosts compatibility-based solvers past PuzzleFlow~(Sec.~\ref{sec:experiments}).

\paragraph{Pretrained visual backbones.}
We build on frozen patch-level features, which encode interior content unaffected by boundary removal.
MAE~\cite{he2022masked} reconstructs masked patches, yielding tokens rich in local texture, whereas backbones pretrained with classification~\cite{dosovitskiy2021image}, self-distillation~\cite{DINOv2}, or language-contrastive~\cite{CLIP2021} objectives favor semantics over fine spatial detail, and convolutional baselines~\cite{liu2022convnet,he2016deep} complete our six-backbone comparison (Fig.~\ref{fig:backbone_comparison}).
And while jigsaw puzzles were used as a motivational task for studying representation learning~\cite{NorooziFavaro2016,JigsawPretextFollowup2019}, here we invert this line of thinking and use pretrained features to solve eroded jigsaws.

\section{Problem Setup and Datasets}
\label{sec:setup}

\paragraph{Puzzles and erosion.}
A square jigsaw puzzle is defined by a source image partitioned into an $N{\times}N$ grid of non-overlapping pieces $\{p_i\}_{i=1}^{N^2}$, each having original size $s{\times}s$ pixels. These pieces are then shuffled and must be reassembled computationally.
In an \emph{eroded} puzzle, $e$ pixels are removed from every side of every piece, reducing it to $(s{-}2e){\times}(s{-}2e)$ pixels and creating a $2e$-pixel gap between true neighbors.
To follow the terminology used by the different benchmarks, on ImageNet $7{\times}7$ we report erosion as the level $e$ per side, while on JPwLEG-3/5  the presentation is based on the resulting $2e$-pixel gap. 
Our main experiments follow the standard regime of known grid size and piece orientation. However, we do evaluate robustness for both missing pieces and unknown orientations (Unknown-orientation results are in the Supp).

\paragraph{Compatibility functions and solvers.}
A CM takes an \emph{anchor} $p_i$, a \emph{candidate} $p_j$, and a spatial relation $d \in \mathcal{D} = \{\mathrm{R}, \mathrm{D}, \mathrm{L}, \mathrm{U}\}$ (Right, Down, Left, Up), and returns a score $\mathrm{CM}(p_i, p_j, d) \in \mathbb{R}$ that orders candidates by their likelihood of being direct neighbors.
The polarity of this ordering varies across measures: dissimilarity-based CMs such as SSD, $L_1$, PBC, MGC, and Edge2Vec assign \emph{lower} scores to better matches, whereas STITCH (Sec.~\ref{sec:method:head}) produces similarity scores in $[-1,1]$ via cosine similarity, where \emph{higher} is better.
A solver utilizes the full CM score matrix and produces an assignment of pieces to grid positions.
When the solver treats the CM as a \emph{replaceable} black-box module, any other CM, including STITCH, can be plugged in unchanged, modulo a sign convention.

\paragraph{Datasets.}
We use several different datasets in our study. \emph{JPwLEG-3} and \emph{JPwLEG-5}~\cite{SD2RL2020} contain $3{\times}3$ puzzles with 48\,px gaps and $5{\times}5$ puzzles with 12\,px gaps, respectively, built from MET artwork images, with random positional jitter of ${\pm}7$\,px that imitates real-world misalignment.
\emph{ImageNet $7{\times}7$} is our own new benchmark of larger, more combinatorially challenging puzzles generated from randomly sampled ImageNet images~\cite{imagenet15russakovsky}, divided into a $7{\times}7=49$ pieces of $100{\times}100$\,px, with no image overlap with JPwLEG.
It is released with erosion levels $e \in \{0,\ldots,6\}$\,px and $k \in \{0, 5, 10, 15, 20\}$ missing pieces, over which we sweep in Sec.~\ref{sec:experiments:robustness}.
\emph{GAP-3} and \mbox{\emph{GAP-5}}~\cite{Shahar_2026_CVPR} extend the same grid topologies ($3{\times}3$ and $5{\times}5$) to irregular, heavily eroded fragments whose shapes are learned from real archaeological fragment statistics over MET images.
Split sizes and per-benchmark statistics are in the Supp.

\paragraph{Metrics.}

We employ and report three metrics.
\emph{Top-1} measures CM quality (independently of any solver). For each query of an anchor piece and relation $d \in \mathcal{D}$, we find the best matching piece according to the CM. The query scores 1 if that match is the correct neighboring piece and 0 otherwise. Top-1 averages these scores over all queries in the test set~\cite{cho2010probabilistic}.

At the solver level, \emph{Abs} is the fraction of pieces placed in their correct grid position~\cite{cho2010probabilistic}, and \emph{Perf} is the fraction of puzzles solved perfectly~\cite{PomeranzBenShahar2011}.

\section{STITCH}
\label{sec:method}

\begin{figure}[!tb]
\centering
\includegraphics[width=\columnwidth]{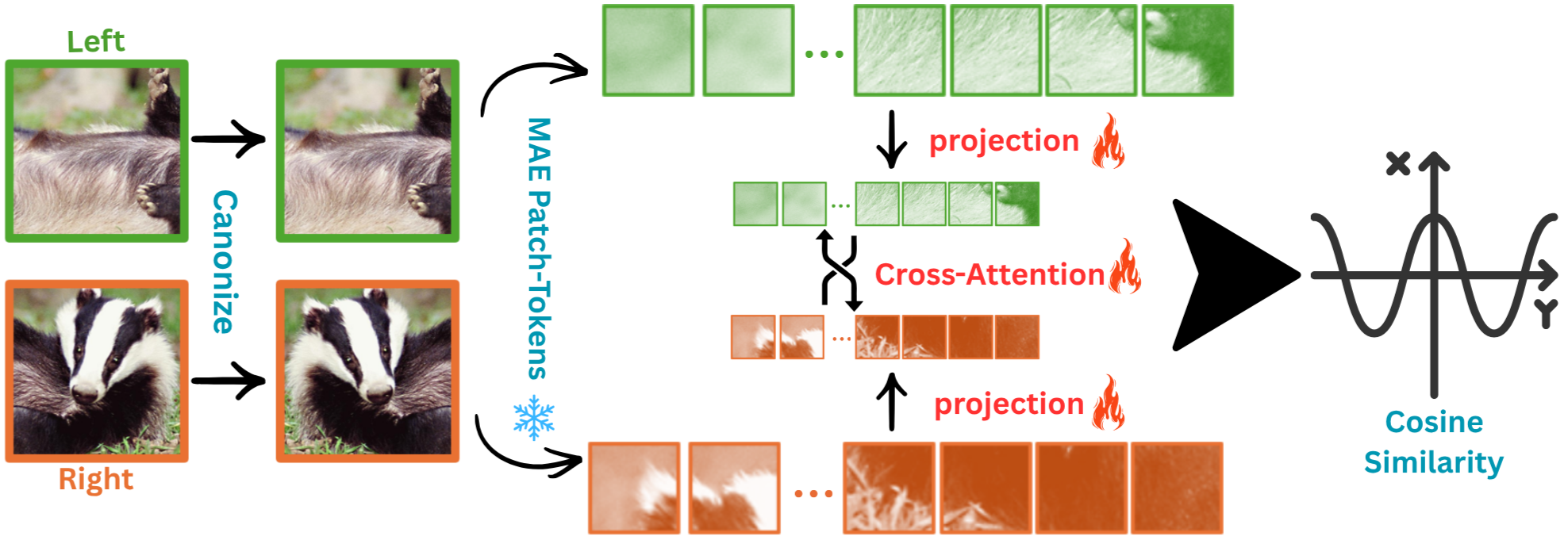}
\caption{%
\textbf{Architecture overview.}
Both pieces are \textcolor{mycyan}{canonicalized} so facing edges align to a fixed position (Sec.~\ref{sec:method:canon}).
A frozen MAE ViT-B/16 extracts \textcolor{mycyan}{$G{\times}G$ patch tokens} (precomputed).
Tokens are \textcolor{myred}{projected} ($768 \to 64$), passed through $L{=}2$ \textcolor{myred}{bidirectional cross-attention} layers (4 heads), and scored via \textcolor{mycyan}{mean tokenwise cosine similarity} ($\sigma \in [-1,1]$).
Only ${\sim}$249\,K parameters are trained.%
}
\label{fig:architecture}
\end{figure}

STITCH is a two-stage pipeline (Fig.~\ref{fig:architecture}): a \emph{frozen} self-supervised ViT extracts patch-level features, and a lightweight \emph{cross-attention head} compares the two feature maps to produce a compatibility score.
Only the head is trained.
The following subsections describe each component.

\subsection{Backbone Feature Extraction}
\label{sec:method:backbone}

Every piece, whatever its size after erosion, is resized to $96{\times}96$\,px before the backbone, so a frozen ViT with patch size $t{=}16$ yields a fixed $G{\times}G$ token map $\mathbf{T}_i \in \mathbb{R}^{G{\times}G{\times}d_{\text{ViT}}}$ with $G{=}6$ and $d_{\text{ViT}}{=}768$, identically across all benchmarks and erosion levels.
Positional embeddings are bicubically interpolated from the backbone's pretraining resolution to this grid.
Crucially, these tokens describe the piece \emph{interior}, the region erosion leaves untouched, rather than its inaccessible eroded boundary.
Features are precomputed and cached, so no ViT forward pass occurs at training time.
We use MAE~\cite{he2022masked}, which consistently outperforms all alternative backbones (Fig.~\ref{fig:backbone_comparison}).

\begin{figure}[!t]
\centering
\includegraphics[width=0.7\columnwidth]{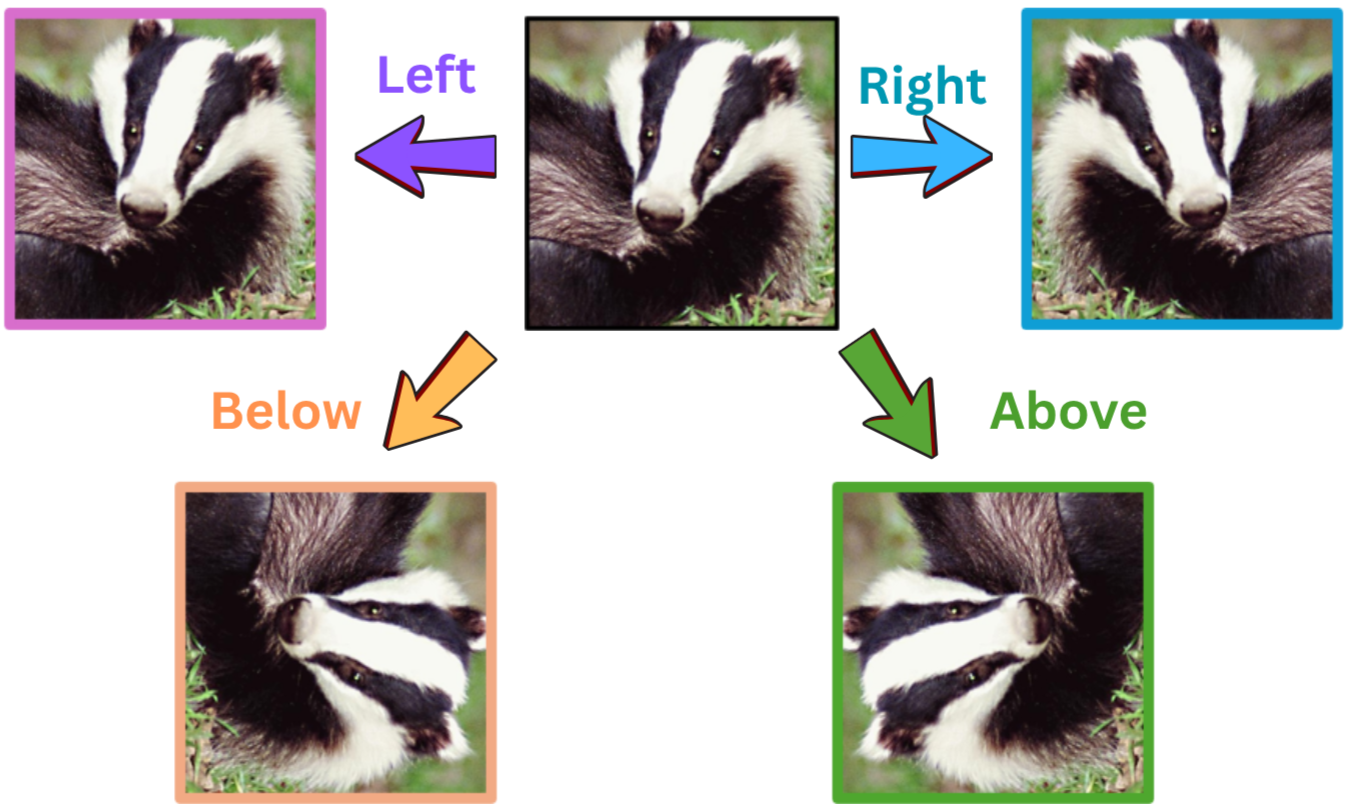}
\caption{%
\textbf{Canonicalization by relative position.}
A piece (center) is transformed according to where it sits relative to
its partner:
\textcolor{Purple}{left} -- identity;
\textcolor{RoyalBlue}{right} -- horizontal flip;
\textcolor{OliveGreen}{above} -- transpose;
\textcolor{Orange}{below} -- $90^{\circ}$ CW rotation.
Each transform brings the piece's facing edge to its rightmost column.%
}
\label{fig:canonicalization}
\end{figure}

\subsection{Canonicalization}
\label{sec:method:canon}

The CM must score candidate neighbors in all four directions $d \in \mathcal{D}$. Rather than learning four separate comparators or conditioning the head on $d$, both of which fragment the training signal, we \emph{canonicalize} the pieces: each piece is transformed by a single deterministic rule that depends only on its position relative to its partner, so that its edge abutting the anchor piece, where the interaction happens, always lands in its rightmost column (Fig.~\ref{fig:canonicalization}). Specifically, a piece \emph{left} of its partner needs no transform (identity) because the interacting edge is already at its right hand side. A piece to the \emph{right} is flipped horizontally, a piece \emph{above} the anchor is transposed, and a piece \emph{below} is rotated $90^{\circ}$~CW. After canonicalization, every pair of anchor and candidate pieces is reduced to the same comparison regardless of their spatial relationship -- they become two pieces whose abutting edges are in the same (right-facing) canonical position.

Formally, let $\rho_r$ denote this transform for relative position
$r \in \mathcal{D}$, and write $\mathbf{T}_i^{d}$ for the map of $p_i$
canonicalized against a partner lying in direction $d$ from it. Since
$p_i$ then lies in the opposite direction $\bar d$ from that partner,
the transform applied is $\rho_{\bar d}$, and symmetrically for the
candidate:
\begin{equation}
    \mathbf{T}_i^d = \rho_{\bar d}(\mathbf{T}_i)\,,\qquad
    \mathbf{T}_j^{\bar d} = \rho_{d}(\mathbf{T}_j)\,.
\end{equation}
This reduces four directional comparisons to a single learned function,
concentrating the training signal in one comparator and keeping the head
small. All four transformed variants per piece are precomputed.
An alternative \emph{dual-canonical} scheme maps horizontal and vertical
edges to separate positions. We compare both and provide details in the Supp.

\subsection{Cross-Attention Interaction Head}
\label{sec:method:head}

Given canonicalized token maps $\mathbf{T}_i^{d}$ and $\mathbf{T}_j^{\bar d}$, each flattened to $G^2$ tokens, the head produces a scalar score.
Tokens are first projected to $d_{\text{model}}{=}64$ via a shared linear layer with layer norm and GELU.
The projected sequences then pass through $L{=}2$ bidirectional cross-attention layers (4 heads), each performing multi-head attention in both directions followed by a feed-forward network with residual connections.
This is where compatibility is estimated: every interior token of the anchor can attend to every interior token of the candidate, letting the head discover texture continuations that no boundary comparison could see.
No additional positional embeddings are added. The frozen ViT tokens already encode spatial structure.

Finally, the two output streams are compared by mean tokenwise cosine similarity:
\begin{equation}
\mathrm{CM}(p_i, p_j, d)
\;=\;
\tfrac{1}{G^2} {\textstyle\sum_{n=1}^{G^2}}
\frac{\mathbf{a}_n \cdot \mathbf{c}_n}
{\|\mathbf{a}_n\|\;\|\mathbf{c}_n\|}
\;\in\; [-1, 1]\,,
\label{eq:cm}
\end{equation}
where $\mathbf{a}_n$ and $\mathbf{c}_n$ are the $n$-th output tokens of the anchor and candidate streams.
The full head contains ${\sim}$249\,K trainable parameters, all in the projection and cross-attention layers.

\subsection{Training}
\label{sec:method:training}

\paragraph{Pair construction and loss.}
For each piece and direction, the true grid neighbor forms a positive pair. Negatives are sampled 50/50 from intra-puzzle (same image, non-adjacent) and cross-puzzle pools ($M{=}32$ per anchor).
Adjacency labels come for free from the known grid, so no manual annotation is required.
Writing $\sigma^+$ for the score of the positive pair and $\sigma_m^-$ for that of the $m$-th negative, we train with InfoNCE~\cite{oord2018representation}:
\begin{equation}
\mathcal{L} = -\log
\frac{\exp(\tau \cdot \sigma^+)}
{\exp(\tau \cdot \sigma^+)
+ {\textstyle\sum_{m=1}^{M}}
\exp(\tau \cdot \sigma_m^-)}\,,
\label{eq:infonce}
\end{equation}
where $\tau$ is a learnable inverse temperature (init.\ $1/0.07$). Label smoothing of $0.05$ is applied, omitted above for clarity.
Each batch uses a single edge direction, so the head cannot key on transform-specific artifacts instead of content continuity.

\paragraph{Optimization.}
All models share identical hyperparameters ($d_{\text{model}}{=}64$, $L{=}2$, 4 heads, ``right column'' canonicalization) across all benchmarks, with no benchmark-specific tuning. The same configuration is reused on JPwLEG-3/5, ImageNet $7{\times}7$, and GAP.
The full optimization protocol (AdamW, schedule, early stopping) is provided in the Supp.

\section{Experiments}
\label{sec:experiments}

\subsection{Experimental Setup}
\label{sec:experiments:setup}

Each CM is trained once per benchmark and then frozen. The same head is
reused across all solvers, erosion levels, and missing-piece counts
without fine-tuning, and on ImageNet $7{\times}7$ training uses
$e{=}4$\,px only.
We compare four handcrafted CMs (SSD~\cite{cho2010probabilistic},
$L_1$~\cite{paikin2015solving}, PBC~\cite{PomeranzBenShahar2011},
MGC~\cite{Gallagher2012}) and our reimplementation of
Edge2Vec~\cite{Edge2Vec2022}, each plugged into three
compatibility-based solvers (greedy~\cite{paikin2015solving},
genetic~\cite{Sholomon2013}, LP~\cite{YuRussellAgapito2016}), negating
dissimilarity CMs so every solver always consumes a higher-is-better CM score.
On ImageNet $7{\times}7$ we add \emph{inpaint-then-match} pipelines that
restore the eroded gap with SDXL~\cite{podell2024sdxl},
LaMa~\cite{suvorov2022resolution}, or MAE inpainting before applying a
classical boundary CM, following \citet{Bridger2020}. On GAP we compare
against PuzzleFlow~\cite{Shahar_2026_CVPR} and its official baselines.
Protocol details are in the Supp.

\begin{figure}[!t]
\centering
\includegraphics[width=0.8\columnwidth]{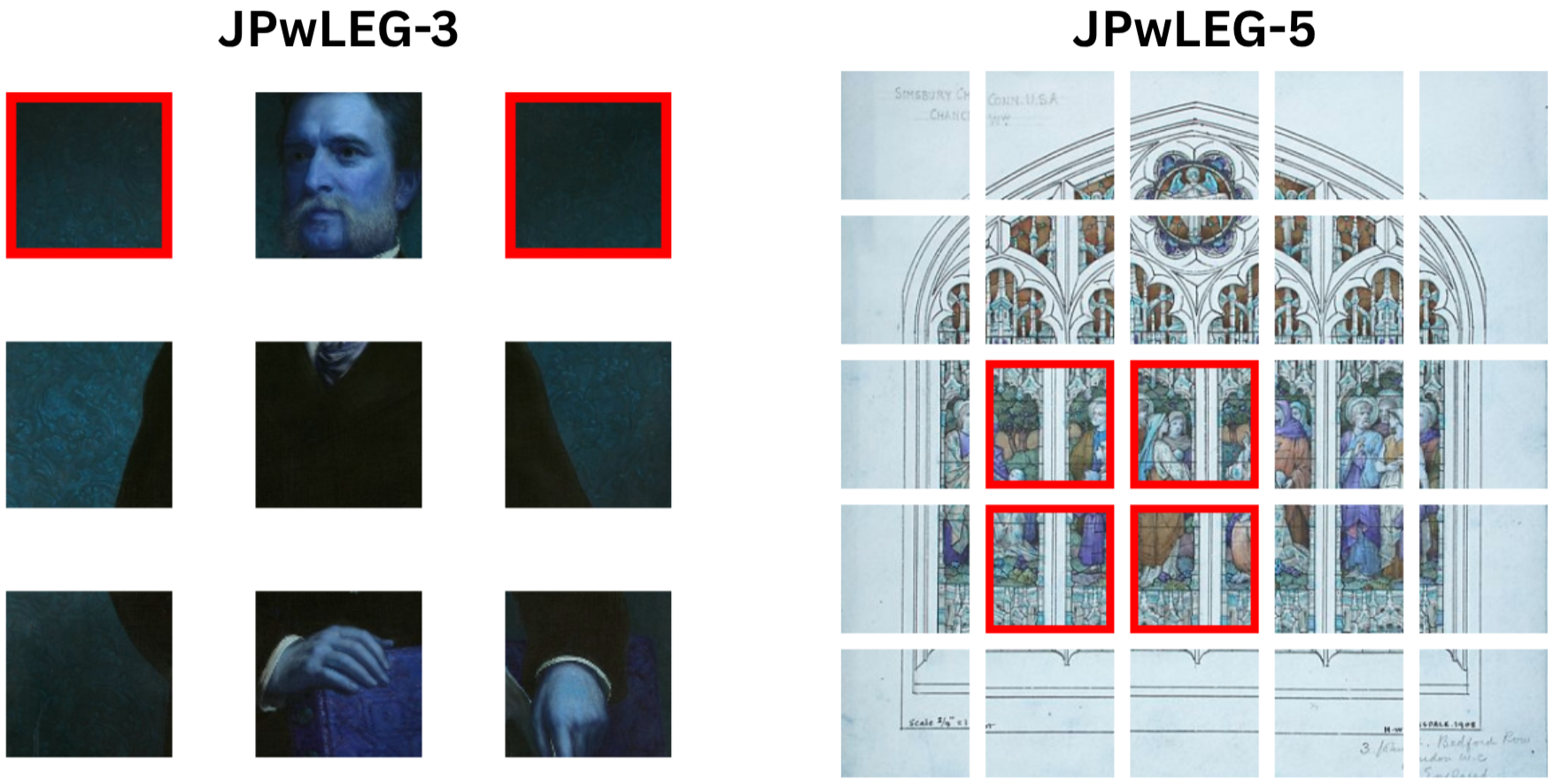}
\caption{%
\textbf{Qualitative results on JPwLEG} (STITCH + genetic solver).
JPwLEG-3 (\emph{left}, 48\,px gap) and JPwLEG-5 (\emph{right}, 12\,px
gap). \textcolor{red}{Red} frames mark misplaced pieces, which occur
only on near-uniform dark pieces (\emph{left}) and repetitive
stained-glass panels (\emph{right}), where interior content is
interchangeable.%
}
\label{fig:jpwleg_viz}
\end{figure}

\subsection{Oracle Analysis and CM Swap on JPwLEG}
\label{sec:experiments:oracle}

\begin{figure}[t]
\centering
\includegraphics[width=0.9\columnwidth]{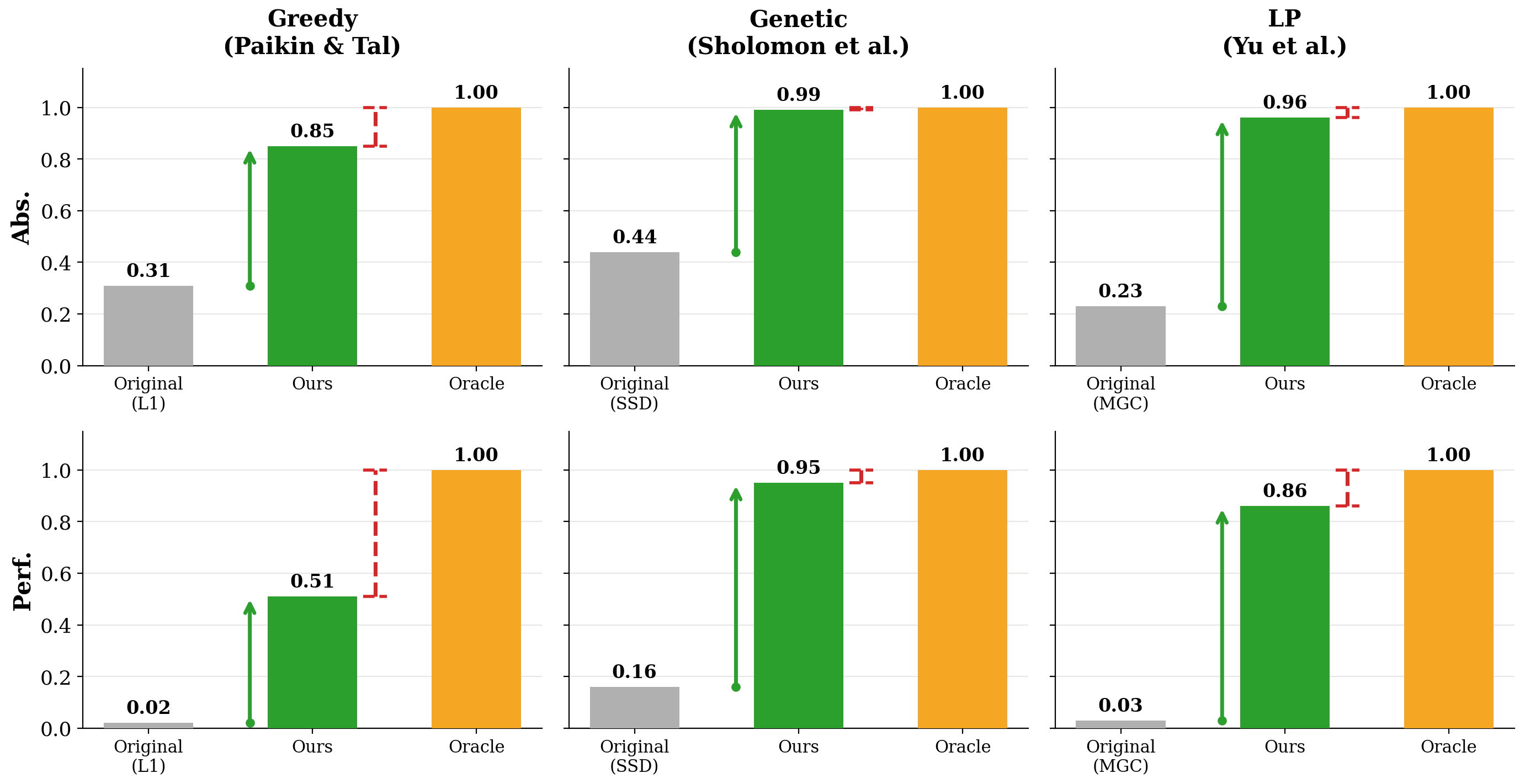}
\caption{%
\textbf{Oracle ceiling (JPwLEG-5).}
Abs (top) and Perf (bottom) for each solver with its original CM (gray),
STITCH (green), and a perfect oracle (gold). All solvers reach 1.00 with
the oracle. STITCH closes most of the gap.%
}
\label{fig:oracle_bottleneck}
\end{figure}

\begin{table*}[!tb]
\centering
\small
\setlength{\tabcolsep}{2pt}
\begin{minipage}[t]{0.48\textwidth}
\centering
\textbf{JPwLEG-3} ($3{\times}3$, 48\,px gap)\\[4pt]
\begin{tabular}{l | c | cc | cc | cc}
\toprule
& & \multicolumn{2}{c}{Greedy}
& \multicolumn{2}{c}{Genetic}
& \multicolumn{2}{c}{LP} \\
\cmidrule(lr){3-4} \cmidrule(lr){5-6} \cmidrule(lr){7-8}
CM & Top-1 & Abs & Perf & Abs & Perf & Abs & Perf \\
\midrule
SSD      & 38.5 & .43 & .09 & .43 & .13 & .47 & .16 \\
$L_1$    & 42.5 & .46 & .11 & .50 & .23 & .54 & .26 \\
PBC      & 36.6 & .41 & .08 & .40 & .11 & .44 & .13 \\
MGC      & 38.1 & .48 & .12 & .45 & .16 & .48 & .19 \\
Edge2Vec & 50.8 & .56 & .17 & .70 & .44 & .74 & .49 \\
\textbf{STITCH}
& \textbf{80.1} & \textbf{.84} & \textbf{.64}
& \textbf{.97} & \textbf{.92} & \textbf{.98} & \textbf{.93} \\
\bottomrule
\end{tabular}
\end{minipage}%
\hfill
\begin{minipage}[t]{0.48\textwidth}
\centering
\textbf{JPwLEG-5} ($5{\times}5$, 12\,px gap)\\[4pt]
\begin{tabular}{l | c | cc | cc | cc}
\toprule
& & \multicolumn{2}{c}{Greedy}
& \multicolumn{2}{c}{Genetic}
& \multicolumn{2}{c}{LP} \\
\cmidrule(lr){3-4} \cmidrule(lr){5-6} \cmidrule(lr){7-8}
CM & Top-1 & Abs & Perf & Abs & Perf & Abs & Perf \\
\midrule
SSD      & 51.3 & .27 & .01 & .44 & .16 & .23 & .05 \\
$L_1$    & 56.3 & .31 & .02 & .56 & .37 & .34 & .11 \\
PBC      & 48.8 & .26 & .01 & .41 & .15 & .22 & .04 \\
MGC      & 52.6 & .30 & .01 & .57 & .27 & .23 & .03 \\
Edge2Vec & 75.0 & .61 & .19 & .93 & .81 & .69 & .46 \\
\textbf{STITCH}
& \textbf{83.5} & \textbf{.85} & \textbf{.51}
& \textbf{.99} & \textbf{.95} & \textbf{.96} & \textbf{.86} \\
\bottomrule
\end{tabular}
\end{minipage}
\caption{%
Averaged over 2\,000 test puzzles and 5 runs, SEM ${\leq}\,0.01$.
All STITCH-vs-baseline gaps are significant at $p<.001$ under paired,
corrected two-sided tests (Supp).%
}
\label{tab:solver_swap}
\end{table*}

We first replace each solver's CM with an oracle returning 1 for true
neighbors and 0 otherwise. Every solver, including the simplest greedy
strategy, achieves perfect Abs and Perf with the oracle
(Fig.~\ref{fig:oracle_bottleneck}): failures on eroded puzzles are
failures of \emph{matching}, not search, the central thesis of this
work.

Table~\ref{tab:solver_swap} then tests how much of this oracle gap
STITCH closes by swapping only the CM. STITCH attains the highest Top-1
on both benchmarks by a wide margin, and the gain propagates through
every solver, lifting the genetic and LP solvers to within a few Abs
points of the oracle ceiling. On JPwLEG-5 the genetic solver rises from
.93 Abs with the strongest prior CM (Edge2Vec) to a near perfect .99, almost doubling
the .57 of the best handcrafted CM (MGC).

\subsection{Comparison with Prior Work}
\label{sec:experiments:sota}

\begin{table}[t]
\centering
\small
\setlength{\tabcolsep}{3.5pt}
\begin{tabular}{l cc cc}
\toprule
& \multicolumn{2}{c}{JPwLEG-3} & \multicolumn{2}{c}{JPwLEG-5} \\
\cmidrule(lr){2-3} \cmidrule(lr){4-5}
Method & Abs & Perf & Abs & Perf \\
\midrule
Deepzzle~\shortcite{Deepzzle2020}    & .738 & .523 & .219 & .000 \\
SD$^2$RL~\shortcite{SD2RL2020}       & .816 & .597 & .403 & .051 \\
PDN-GA~\shortcite{PDNGA2021}         & .813 & .582 & .443 & .061 \\
JPDVT~\shortcite{JPDVT2023}          & .713 & --   & .041 & .000 \\
ERL-MPP~\shortcite{song2025erl}      & --   & --   & .527 & .186 \\
FCViT~\shortcite{kim2025solving}     & \underline{.969} & \underline{.879} & .432 & .002 \\
DiffAssemble~\shortcite{scarpellini2024diffassemble} & .797 & .542 & .629 & .164 \\
PuzLM~\shortcite{elkin2025seq2seq}   & .919 & .845 & \underline{.721} & \underline{.325} \\
\midrule
\textbf{STITCH} (genetic solver)
& \textbf{.973} & \textbf{.925} & \textbf{.992} & \textbf{.954} \\
\bottomrule
\end{tabular}
\caption{%
\textbf{Comparison with prior work on JPwLEG.}
Best in \textbf{bold}, second-best \underline{underlined}, unavailable
entries \mbox{--}.
All prior methods are learning-based solvers. Ours is a modular CM
plugged into a genetic solver~\cite{Sholomon2013}.%
}
\label{tab:sota}
\end{table}

Pairing STITCH with the genetic solver, the strongest on JPwLEG-5
(Table~\ref{tab:solver_swap}), surpasses all learning-based solvers on
both benchmarks (Table~\ref{tab:sota}). The margin over the runner-up is
narrow on JPwLEG-3 but widens dramatically on JPwLEG-5 (+.629 Perf over
PuzLM), leaving virtually no room for further gains. Since
Table~\ref{tab:solver_swap} shows the same head upgrading \emph{every}
solver, better matching, not better search, drives this leap.

\subsection{Upscaling performance using ImageNet $7{\times}7$: robustness to erosion level and missing pieces}

\label{sec:experiments:robustness}

\begin{figure}[t]
\centering
\includegraphics[width=0.8\columnwidth]{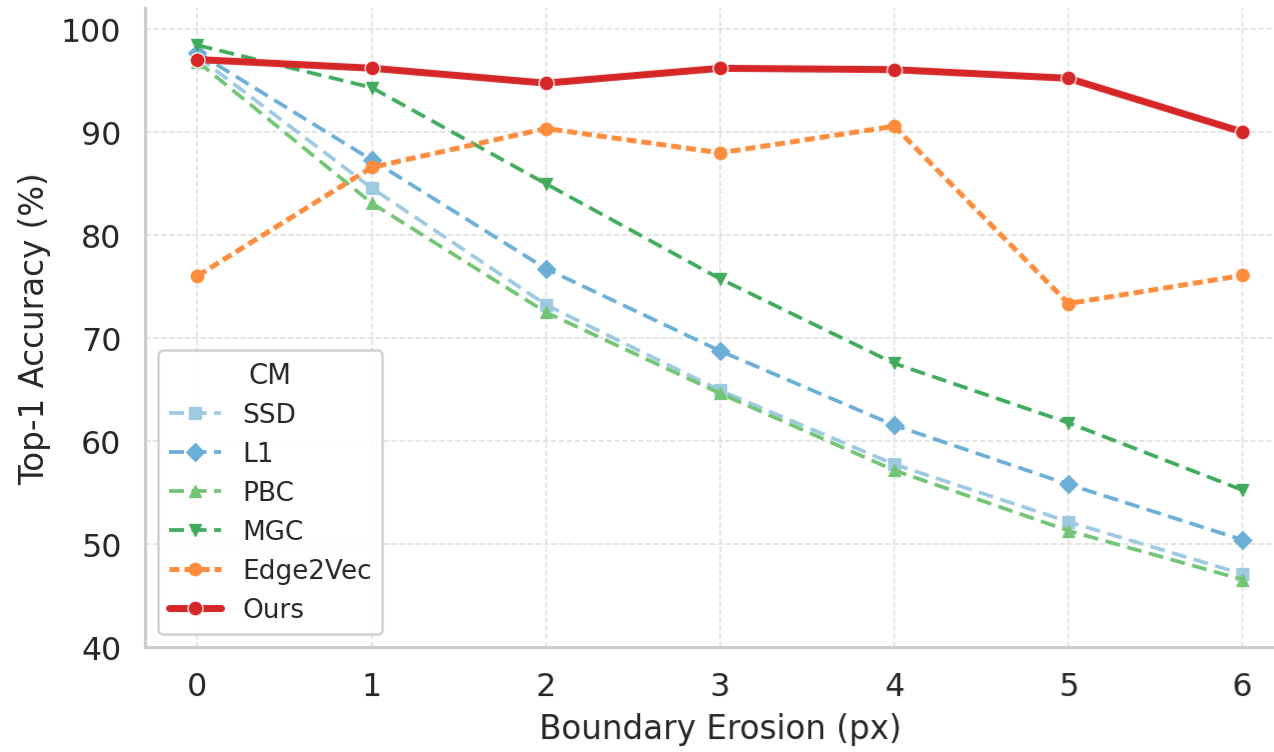}
\caption{%
\textbf{Erosion robustness (ImageNet $7{\times}7$).}
Top-1 vs.\ erosion (px). Learned CMs (STITCH, Edge2Vec) trained once at
4\,px. STITCH (red) remains nearly flat while all baselines degrade
sharply.%
}
\label{fig:erosion_sweep}
\end{figure}

\begin{table}[t]
\centering
\small
\setlength{\tabcolsep}{3pt}
\begin{tabular}{c cc cc cc}
\toprule
& \multicolumn{2}{c}{MGC} & \multicolumn{2}{c}{Edge2Vec}
& \multicolumn{2}{c}{\textbf{STITCH}} \\
\cmidrule(lr){2-3} \cmidrule(lr){4-5} \cmidrule(lr){6-7}
$k$ & Abs & Perf & Abs & Perf & Abs & Perf \\
\midrule
 0 & .42 & .12 & .91 & .78 & \textbf{.97} & \textbf{.91} \\
 5 & .27 & .00 & .85 & .39 & \textbf{.97} & \textbf{.82} \\
10 & .21 & .00 & .76 & .10 & \textbf{.94} & \textbf{.66} \\
15 & .18 & .00 & .66 & .03 & \textbf{.89} & \textbf{.54} \\
20 & .18 & .00 & .62 & .01 & \textbf{.87} & \textbf{.53} \\
\bottomrule
\end{tabular}
\caption{%
\textbf{Missing pieces} (ImageNet $7{\times}7$, LP solver, 4\,px
erosion). $k$ pieces removed at test time, CMs \emph{not} retrained.
Best per row in \textbf{bold}.%
}
\label{tab:missing_pieces}
\end{table}

ImageNet $7{\times}7$ scales to 49 pieces while varying erosion and missing pieces independently, which JPwLEG's fixed per-grid erosion cannot support.
Across erosion levels \mbox{$e \in \{0,\ldots,6\}$\,px,} STITCH's Top-1 stays nearly flat (see Fig.~\ref{fig:erosion_sweep}), dropping only ${\sim}$8\,pt across the full range, while handcrafted CMs collapse by 40--50\,pt from their clean-boundary peak and Edge2Vec degrades sharply beyond its training erosion.
A natural reservation is that modern generative models would make handling erosion easy: inpaint the gap, then match boundaries. Unfortunately, they do not. At $e{=}6$\,px, restoring pieces to their original $100{\times}100$\,px before applying the best classical CM (MGC) reaches 48.90 Top-1 with SDXL, 56.85 with
LaMa, and 66.08 with MAE inpainting, against 90.00 for STITCH when applied directly on the eroded pieces. Generative inpainting hallucinates plausible but unfaithful content precisely where matching needs evidence, and costs a full generative forward pass per candidate pair. STITCH consults the intact interior instead. Full grids over four CMs are in the Supp.

We further removed $k \in \{0,5,10,15,20\}$ pieces at test time (4\,px erosion, no retraining), choosing removed pieces randomly but in a way that preserves a single connected block of pieces rather than scattered "islands". Table~\ref{tab:missing_pieces} reports the performance of the LP solver, which allows unoccupied positions.
STITCH degrades gracefully even when over 40\% of the pieces are removed.
At $k{=}20$, learning-based solvers trained \emph{with} missing pieces reach .68/.07 Abs/Perf (DiffAssemble) and .02/.00 (JPDVT), against STITCH's .87/.53 without any missing-piece training (full tables, solver$\times$CM grids, and unconstrained removal are in the Supp).

\subsection{Irregular Fragments: GAP-3/5}
\label{sec:experiments:gap}

\begin{figure}[!t]
\centering
\includegraphics[width=0.8\columnwidth]{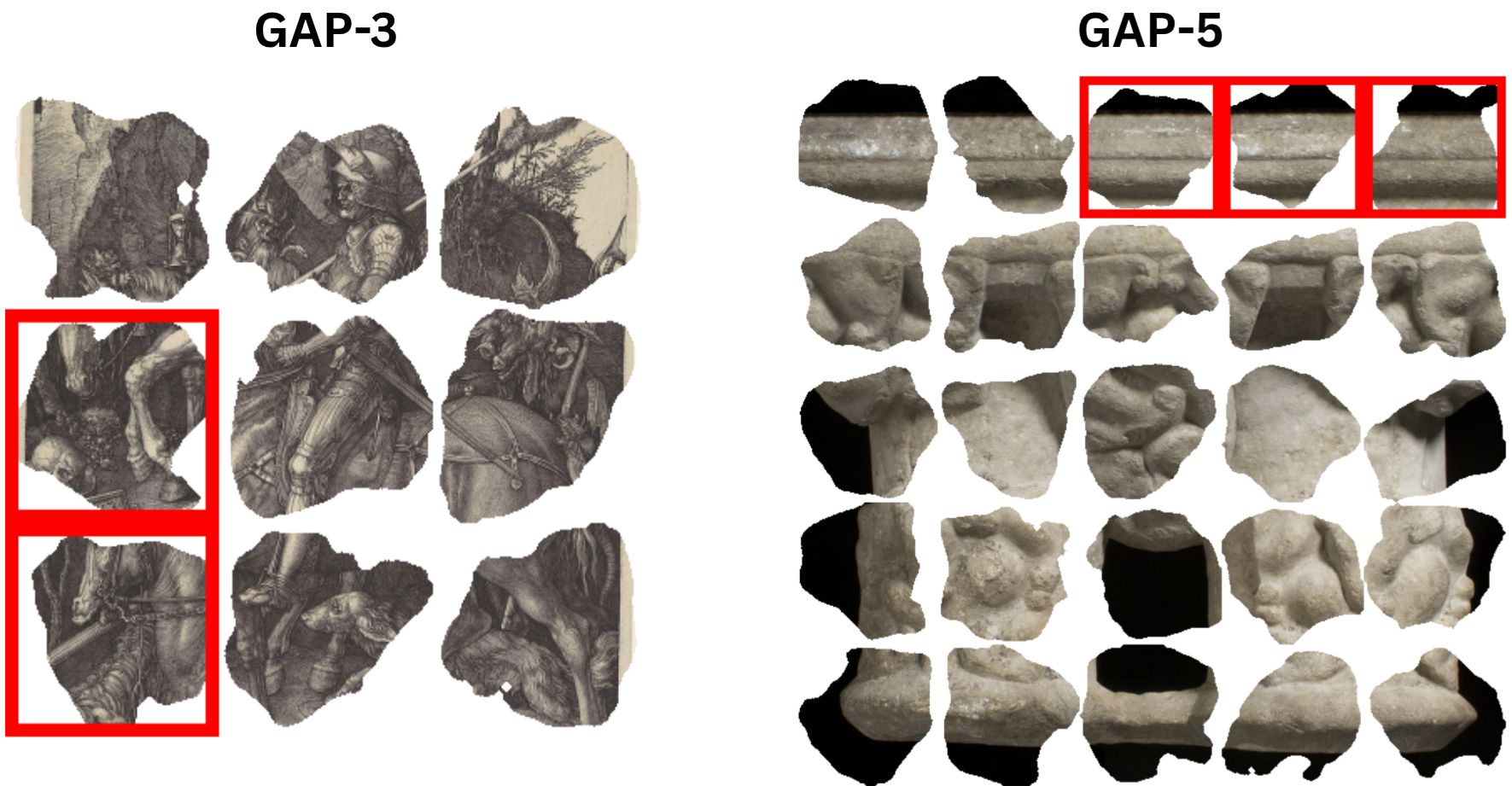}
\caption{%
\textbf{Qualitative results on GAP} (STITCH + genetic solver).
GAP-3 (\emph{left}) and GAP-5 (\emph{right}).
\textcolor{red}{Red} frames mark misplaced fragments. All others are
placed correctly despite the irregular, heavily eroded shapes.%
}
\label{fig:gap_viz}
\end{figure}

\begin{table}[t]
\centering
\small
\setlength{\tabcolsep}{3pt}
\begin{tabular}{l cc cc}
\toprule
& \multicolumn{2}{c}{GAP-3} & \multicolumn{2}{c}{GAP-5} \\
\cmidrule(lr){2-3} \cmidrule(lr){4-5}
Method & Abs & Perf & Abs & Perf \\
\midrule
JigsawGAN~\shortcite{li2021jigsawgan}       & .453 & .046 & .180 & .000 \\
DiffAssemble~\shortcite{scarpellini2024diffassemble} & .505 & .164 & .219 & .000 \\
JPDVT~\shortcite{JPDVT2023}                 & .112 & .000 & .039 & .000 \\
PuzLM~\shortcite{elkin2025seq2seq}          & .148 & .000 & .078 & .000 \\
FCViT~\shortcite{kim2025solving}            & .607 & .252 & .204 & .000 \\
PuzzleFlow~\shortcite{Shahar_2026_CVPR}     & .629 & .285 & .291 & .003 \\
\midrule
\textbf{STITCH} (genetic) & \textbf{.891} & \textbf{.810} & \textbf{.667} & \textbf{.472} \\
\textbf{STITCH} (LP)      & \underline{.883} & \underline{.795} & \underline{.657} & \underline{.262} \\
\bottomrule
\end{tabular}
\caption{%
\textbf{Irregular fragments (GAP-3/5).}
STITCH is trained on GAP~\shortcite{Shahar_2026_CVPR} and plugged into
unchanged solvers. Baselines as originally published, with two accuracy
measures matching Abs and Perf. Best in \textbf{bold}, second-best
\underline{underlined}, \mbox{--} unavailable. Greedy-solver, Top-1, and
neighbor accuracy in the Supp.%
}
\label{tab:gap}
\end{table}

Finally, we tested whether our compatibility survives irregular geometry, the regime PuzzleFlow~\cite{Shahar_2026_CVPR} addresses by abandoning pairwise compatibility altogether. We train STITCH on the GAP training split, alpha-compositing each RGBA fragment over a constant black background into a fixed square before the frozen MAE, so the irregular shape enters through the pixels alone with no shape-aware change to the head. Plugging the trained STITCH into the genetic and LP solvers surpasses every learning-based solver on GAP-3, clearing the previous best by over .25 Abs and .50 Perf. The genetic solver with STITCH perform similarly on GAP-5 as well (Table~\ref{tab:gap}): it more than doubles PuzzleFlow's Abs and raises Perf from .003 to .472, so nearly half of all GAP-5 puzzles are now reconstructed perfectly (Fig.~\ref{fig:gap_viz}). The pairwise-CM paradigm thus extends to irregular fragments once the match itself is fixed.

\subsection{Qualitative Results and Computational Cost}
\label{sec:experiments:qualitative}

\begin{figure}[!t]
\centering
\includegraphics[width=0.8\columnwidth]{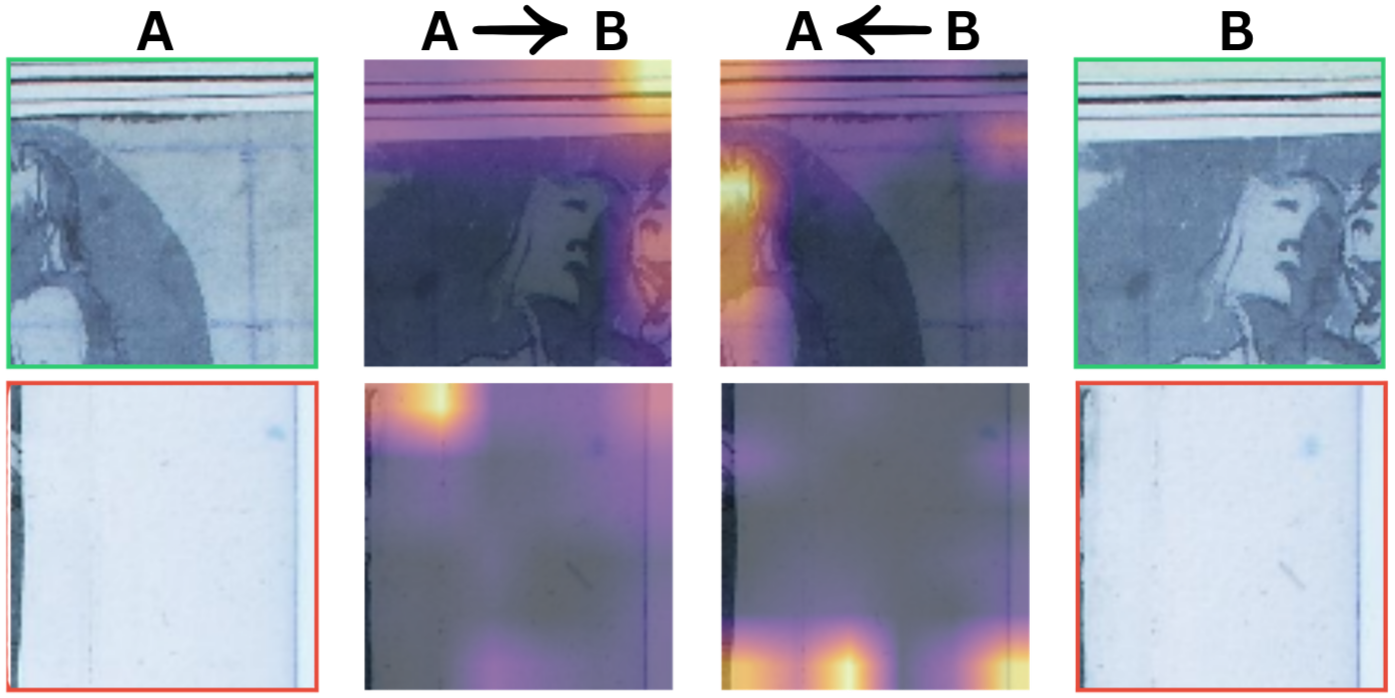}
\caption{%
\textbf{Cross-attention on JPwLEG-5.}
Outer columns show anchor A and candidate B, inner columns the attention
between them (A$\to$B and A$\gets$B), aggregated over all heads and query
tokens in the last layer.
\emph{Top} (\textcolor{ForestGreen}{green}): on pieces with rich
pictorial content, attention concentrates on the facing edges and STITCH
ranks B first.
\emph{Bottom} (\textcolor{red}{red}): on near-textureless pieces,
attention finds no structure to key on and the true neighbor is missed.%
}
\label{fig:attention_viz}
\end{figure}

While presenting a leap from the prior art on compatibility, STITCH is not perfect. The few remaining errors, however, concentrate in textureless or repetitive regions where interior tokens become interchangeable, and compatibility is arguably undecidable (Figs.~\ref{fig:jpwleg_viz} and~\ref{fig:gap_viz}): near-uniform dark pieces in JPwLEG-3, recurring stained-glass panels in JPwLEG-5, and sparse or repetitive fragments in GAP. Fig.~\ref{fig:attention_viz} suggests why this happens. When pieces carry pictorial content, cross-attention anchors onto the facing edges and the true neighbor is ranked first, whereas on textureless pieces attention finds nothing to lock on and the match fails, consistent with a head that has learned spatially meaningful, facing-edge-driven patterns. Additional failure examples are in the Supp.

Computationally, scoring one pair costs a single ${\sim}$23\,MFLOP head forward pass over cached ViT features, so all 4{,}704 pairs of a $7{\times}7$ puzzle are scored in under 0.1\,s on one RTX~4090 (full hardware and software specifications in the Supp), making the solver, not the CM, the runtime bottleneck. Feature extraction is a one-time cost of 196 backbone passes per puzzle, amortized across every pair. Learning-based solvers of comparable accuracy carry 85--124M \emph{trained} parameters~\cite{Shahar_2026_CVPR}, whereas STITCH trains a relatively negligible ${\sim}$249\,K parameters on top of an off-the-shelf frozen backbone.

\section{Ablations}
\label{sec:ablations}

\begin{figure}[!t]
\centering
\includegraphics[width=0.90\columnwidth]{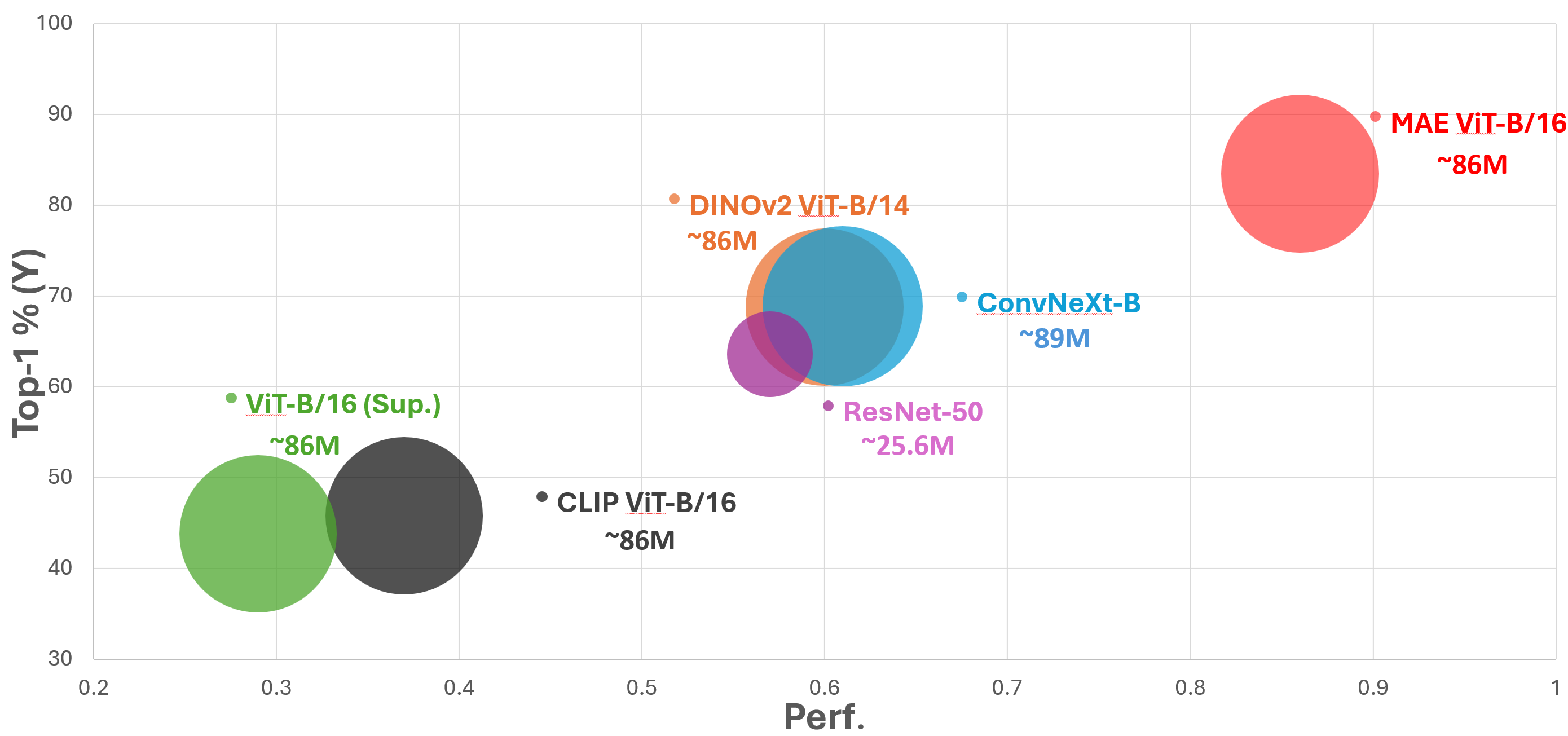}
\caption{%
\textbf{Backbone comparison} (JPwLEG-5, LP solver).
Six frozen backbones positioned by solver-level Perf ($x$) and CM-level
Top-1 ($y$), with area proportional to parameter count.
\textcolor{BubbleRed}{MAE ViT-B/16} dominates both axes despite matching
\textcolor{BubbleOrange}{DINOv2}, \textcolor{BubbleGray}{CLIP}, and
supervised \textcolor{BubbleGreen}{ViT-B/16} in size (${\sim}$86M) and
architecture: the objective, not capacity, drives performance.%
}
\label{fig:backbone_comparison}
\end{figure}

We ablate the backbone and the head configuration, both on JPwLEG-5 with
the LP solver.

\paragraph{Backbone.}
Fig.~\ref{fig:backbone_comparison} compares six frozen backbones.
MAE ViT-B/16 occupies the top-right corner alone, and the three
backbones that share its architecture and size, differing only in
pretraining objective, trail it by up to 39.7 Top-1 points:
classification, self-distillation, and contrastive-language pretraining
all discard the local texture signal that eroded matching requires,
whereas MAE's reconstruction objective preserves it.
A ResNet-50 with a third of the parameters outscores two of them,
confirming that the objective, not capacity or architecture, is what
matters.

\paragraph{Head configuration.}
Table~\ref{tab:ablation} varies the two head hyperparameters around the
default ($d_{\text{model}}{=}64$, $L{=}2$, marked $\dagger$).
Cross-attention is essential: removing it ($L{=}0$) collapses Perf from
.868 to .240. One layer recovers most of the gain, and depth beyond two
adds ${\leq}1$ Top-1 point at nearly double the parameters.
For the projection, $d_{\text{model}}{=}16$ loses over 6 Top-1 points,
$d_{\text{model}}{=}128$ only matches the default, and no projection
costs two orders of magnitude more compute while \emph{reducing}
accuracy.
We therefore adopt the cheapest configuration on the accuracy plateau.
Full sweeps are in the Supp.

\begin{table}[!t]
\centering
\small
\setlength{\tabcolsep}{3.5pt}
\begin{tabular}{l cc ccc}
\toprule
Variant & Params & MFLOPs & Top-1 & Abs & Perf \\
\midrule
\multicolumn{6}{l}{\emph{Projection dimension $d_{\text{model}}$
($L{=}2$)}} \\
$d_{\text{model}}{=}16$          & ${\sim}$25\,K  & 3  & 77.2 & .890 & .742 \\
$d_{\text{model}}{=}64^\dagger$  & ${\sim}$249\,K & 23 & 83.5 & .964 & .868 \\
$d_{\text{model}}{=}128$         & ${\sim}$891\,K & 73 & 83.5 & .965 & .867 \\
$d_{\text{model}}{=}d_{\text{ViT}}$ (no proj.) & ${\sim}$28.3\,M & 2{,}139 & 80.1 & .948 & .834 \\
\midrule
\multicolumn{6}{l}{\emph{Cross-attention layers $L$
($d_{\text{model}}{=}64$)}} \\
$L{=}0$ (no cross-attn.) & ${\sim}$49\,K  & 7  & 64.8 & .522 & .240 \\
$L{=}1$                  & ${\sim}$149\,K & 15 & 82.5 & .950 & .848 \\
$L{=}2^\dagger$          & ${\sim}$249\,K & 23 & 83.5 & .964 & .868 \\
$L{=}4$                  & ${\sim}$449\,K & 38 & 84.5 & .967 & .873 \\
\bottomrule
\end{tabular}
\caption{%
\textbf{Head configuration ablation} (JPwLEG-5, LP solver).
Each block varies one hyperparameter with the other fixed at the
default. $\dagger$ marks the chosen configuration, the cheapest on the
accuracy plateau.%
}
\label{tab:ablation}
\end{table}

\section{Conclusion}
\label{sec:conclusion}

We demonstrate that the bottleneck in puzzle solving is not the solver but the match (namely the compatibility measure), and argue that this component deserves far more attention than it has received.
Toward this end, we introduce STITCH, a pluggable data-driven CM that improves square-puzzle solving regardless of the reconstruction method: training only ${\sim}$249\,K parameters on top of a frozen backbone, and plugging into any solver as-is, it nearly saturates both JPwLEG benchmarks, generalizes to unseen erosion levels and missing pieces without retraining, and, once trained on GAP, boosts compatibility-based solvers to outperform the best learning-based model when solving grid puzzles of irregular fragments.
Natural future work includes extensions to real fragment collections beyond grid topologies~\cite{harel2021crossing,tsesmelis2024re,elkin2025recognizing} and large puzzles where caching the complete compatibility table is impractical.

\bibliography{aaai2027}
\end{document}